\documentclass[letterpaper, 10 pt, conference]{ieeeconf}  % Comment this line out if you need a4paper

\IEEEoverridecommandlockouts                              % This command is only needed if 
\usepackage{graphics} % for pdf, bitmapped graphics files
\usepackage{epsfig} % for postscript graphics files
\usepackage{mathptmx} % assumes new font selection scheme installed
\usepackage{times} % assumes new font selection scheme installed
\usepackage{amsmath} % assumes amsmath package installed
\usepackage{amssymb}  % assumes amsmath package installed
\usepackage{booktabs}
\usepackage{caption}
\usepackage{url}
\usepackage{color, colortbl}
\title{\LARGE \bf
GeoAAC: Geometry-Based Adaptive Action Chunking from Denoising Trajectories in VLA Policies
}

\author{
Xin Chen$^{1}$ \quad
Sen Chen$^{1}$ \quad
Yujuan Ding$^{2}$ \quad
Jian Liu$^{1}$ \quad
Guoqing Wang$^{3}$\\
Wei Ye$^{1}$ \quad
Heng Tao Shen$^{1}$ \quad
Yi Bin$^{1\dagger}$\\[1.5mm]
$^{1}$Tongji University
\qquad
$^{2}$The Hong Kong Polytechnic University\\
$^{3}$University of Electronic Science and Technology of China%
\thanks{$^{\dagger}$Corresponding author.}
}

\begin{document}

\maketitle
\thispagestyle{empty}
\pagestyle{empty}

%%%%%%%%%%%%%%%%%%%%%%%%%%%%%%%%%%%%%%%%%%%%%%%%%%%%%%%%%%%%%%%%%%%%%%%%%%%%%%%%

\begin{abstract}

Action chunking is widely used for action generation and execution in Vision-Language-Action (VLA) policies, yet existing approaches commonly use a fixed action horizon. During a rollout, different task stages may require different levels of action continuity, control precision, and closed-loop feedback, making a fixed horizon unable to accommodate changing control requirements. We propose \textbf{GeoAAC}, a geometry-based adaptive action chunking method for flow-based VLA policies that adjusts the action horizon according to the reliability of the current action prediction. We show that the geometry of Flow Matching denoising trajectories provides process-level information for characterizing prediction reliability, with geometric variation across action prefixes remaining positively correlated with predictive uncertainty. GeoAAC uses this prefix-wise geometry to construct a horizon-wise geometric profile and adaptively determine the action horizon from a single generation without additional training. Experiments with GR00T N1.5 and $\pi_{0.5}$ on LIBERO, LIBERO-Pro, RoboCasa365, and real-world manipulation tasks show consistent improvements over fixed-action-horizon baselines and existing adaptive methods, including up to 8.7 percentage points in simulation and an increase in average real-world success rate from 53.3\% to 74.4\%.

\end{abstract}

%%%%%%%%%%%%%%%%%%%%%%%%%%%%%%%%%%%%%%%%%%%%%%%%%%%%%%%%%%%%%%%%%%%%%%%%%%%%%%%%

\section{INTRODUCTION}

Action chunking has become a widely adopted action-generation and execution strategy in Vision-Language-Action (VLA) policies, which leverage large-scale robotic data and pretrained vision-language representations to generalize across tasks, environments, and embodiments~\cite{brohan2023rt1,zitkovich2023rt2,kim2025openvla,mees2024octo,black2025pi0,black2025pi05,QuD-RSS-25,BuQ-RSS-25,fan2025longvla}. Rather than predicting a single control command at each step, action chunking predicts a sequence of future actions, capturing temporal dependencies and improving motion continuity~\cite{zhao2023act,chi2023diffusionpolicy,liu2025bidirectional,lee2025interact}.

\begin{figure}[t]
    \centering
    \includegraphics[width=\columnwidth]{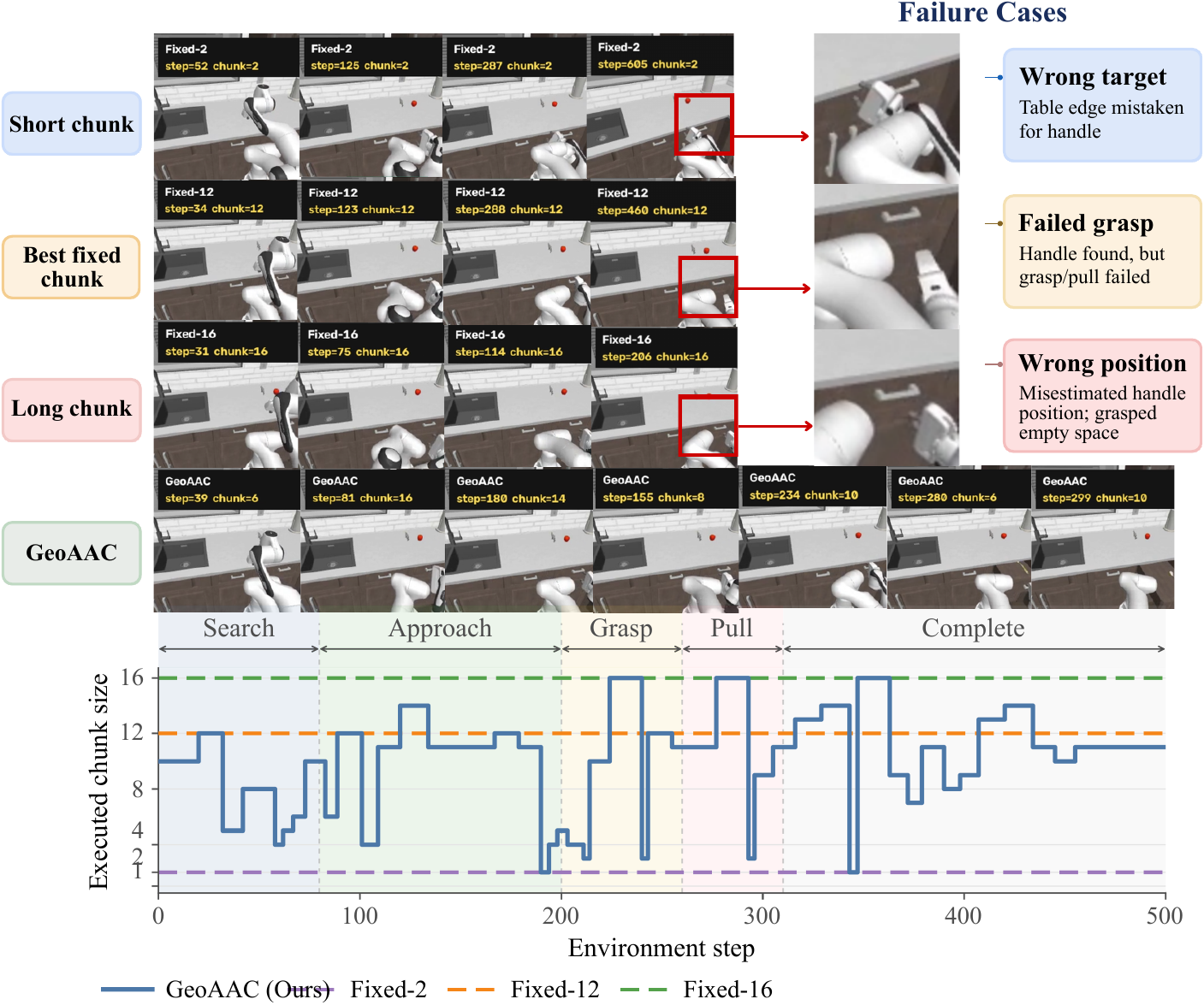}
    \caption{
    \textbf{Stage-dependent action horizon requirements.}
    A representative OpenDrawer rollout in RoboCasa.
    Fixed-$2$ misidentifies the table edge as the handle during search;
    Fixed-$12$ localizes the handle but fails during subsequent grasping and pulling;
    Fixed-$16$ executes a long open-loop segment while approaching the handle and ultimately misses the grasp.
    GeoAAC instead adjusts its action horizon as the rollout progresses.
    }
    \label{fig:stage_dependent_horizon}
\end{figure}

Applying action chunking requires specifying the \emph{action horizon}, i.e., the number of predicted actions executed before the policy observes and replans. The action horizon governs the trade-off between action continuity and closed-loop responsiveness: shorter horizons enable more frequent feedback and correction but require more frequent policy inference, whereas longer horizons preserve motion continuity at the cost of extended open-loop execution~\cite{chi2023diffusionpolicy,liu2025bidirectional,black2025realtime,xue2025reactive,jiang2025streaming}. Most existing VLA policies use a fixed action horizon during deployment~\cite{black2025pi0,black2025pi05,kim2025finetuning}. In practice, however, the appropriate horizon is inherently task- and stage-dependent, as different tasks and stages require different levels of action continuity, control precision, and closed-loop feedback. Consequently, a fixed horizon may suit certain stages while compromising others~\cite{jing2025mixture,zhao2026horizon,liang2026adaptive,nie2026pace,wang2026vlaknows}. As illustrated in Fig.~\ref{fig:stage_dependent_horizon}, representative fixed action horizons fail at different stages of the same manipulation task, highlighting the need to adapt the action horizon as control requirements change throughout a rollout.

To address the limitations of fixed action horizons, recent studies have explored adaptive action chunking, which adjusts the action horizon throughout a rollout to enable timely observation and replanning when the current action prediction becomes less reliable. Some approaches train additional horizon predictors or horizon-specific policies to select the execution boundary from the current observation or task state~\cite{zhao2026horizon,jing2025mixture}, but require additional data, supervision, or optimization. Training-free methods instead seek measurable proxies from the current policy inference to characterize how reliably the predicted action chunk can be executed and thereby determine the action horizon~\cite{liang2026adaptive,wang2026vlaknows,nie2026pace}.
One representative approach estimates predictive uncertainty from multiple stochastic action predictions using action entropy and uses it to guide action horizon selection~\cite{liang2026adaptive}. While such uncertainty estimates indicate prediction reliability, robot action distributions are often multimodal, where multiple distinct action sequences can all represent valid behaviors under the same observation. As a result, output-level discrepancies among sampled predictions may reflect diversity across valid action modes rather than unreliability of the current prediction itself. This motivates us to examine whether the action-generation process itself contains richer reliability information that can better support action horizon selection.

Recent VLA policies employ Flow Matching for continuous action generation~\cite{black2025pi0,black2025pi05,bjorck2025gr00t}. Flow Matching progressively transports initial noise toward the final action output through a sequence of intermediate velocity predictions~\cite{lipman2023flow}. These intermediate predictions capture how the current action prediction is progressively updated and refined throughout the generation process. Prior studies have analyzed this process from a geometric perspective and shown that variations in the velocity field and the resulting trajectory geometry can reflect predictive uncertainty~\cite{liu2023rectifiedflow,pooladian2023multisample,rao2026geometry}: smaller geometric variations correspond to more consistent generative dynamics, whereas larger variations indicate stronger internal adjustments to the current prediction. Accordingly, denoising-trajectory geometry provides process-level information that can more directly characterize the reliability of the current prediction. Building on this observation, we analyze the denoising-trajectory geometry across different action prefixes and find that prefix-wise geometric variation remains positively correlated with predictive uncertainty, as shown in Fig.~\ref{fig:prefix_accel_proxy}. This indicates that Flow Matching geometry provides reliability-related information along the action horizon.

\begin{figure*}[t]
    \centering
    \includegraphics[width=0.8\textwidth]{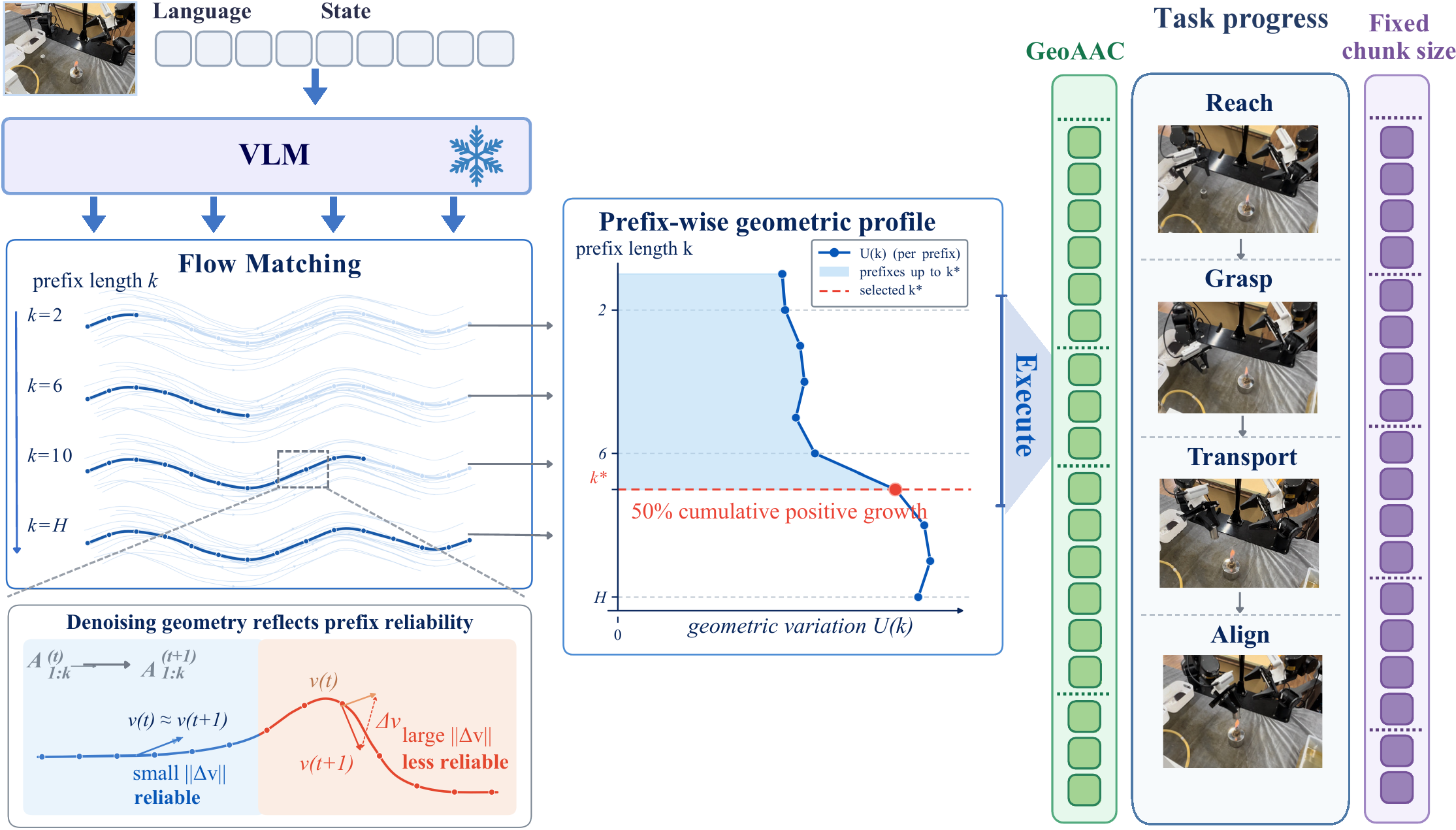}
    \caption{
    \textbf{Overview of GeoAAC.}
    GeoAAC uses the geometry of Flow Matching denoising trajectories to characterize the reliability of different action prefixes and adapt the action horizon accordingly.
    }
    \label{fig:overview}
\end{figure*}

Based on the above observations, we propose \textbf{GeoAAC}, a geometry-based adaptive action chunking method for flow-based VLA policies, as illustrated in Fig.~\ref{fig:overview}. By examining how denoising-trajectory geometry evolves across action prefixes, GeoAAC characterizes how the reliability of the current action prediction changes along the action horizon and uses this information to determine an appropriate execution boundary. Specifically, we first extract local geometric variations for different action prefixes from a single Flow Matching generation. Since the relationship between geometric variation and predictive uncertainty differs across denoising stages, we apply temporal correction and aggregate the stage-wise information into a horizon-wise geometric profile. GeoAAC then determines the execution boundary from the relative growth and cumulative trend of this profile along the action horizon, enabling adaptive action horizon selection according to the reliability of the current prediction without additional training.
We evaluate GeoAAC across multiple flow-based VLA policies and robot manipulation benchmarks. On LIBERO, GeoAAC achieves average success rates of 95.5\% and 98.0\% with GR00T N1.5 and $\pi_{0.5}$, respectively, while outperforming the best fixed-action-horizon baselines by 8.7 and 5.3 percentage points on RoboCasa365 and LIBERO-Pro, respectively. Across three real-world manipulation tasks, GeoAAC further improves the average success rate from 53.3\% with fixed-action-horizon execution to 74.4\%.

Our contributions are summarized as follows:

\begin{itemize}

\item We analyze Flow Matching denoising-trajectory geometry across action prefixes and show that prefix-wise geometric variation remains positively correlated with predictive uncertainty, revealing denoising geometry as a process-level signal of the reliability of the current action prediction along the action horizon.

\item We propose \textbf{GeoAAC}, a geometry-based adaptive action chunking method that aggregates temporally corrected prefix-wise denoising geometry into a horizon-wise geometric profile and selects the action horizon from its cumulative relative growth. GeoAAC enables adaptive action horizon selection from a single Flow Matching generation without additional training.

\item We evaluate GeoAAC with two flow-based VLA policies across three simulation benchmarks and real-world manipulation tasks. GeoAAC consistently outperforms fixed-action-horizon baselines and the compared adaptive action chunking methods.

\end{itemize}

\section{RELATED WORK}

\subsection{Adaptive Action Chunking}

Action chunking is widely used in modern robot policies to generate and execute sequences of future actions
\cite{zhao2023act,chi2023diffusionpolicy,lee2025interact,black2025pi0,black2025realtime,kim2025finetuning}.
While most approaches use a fixed action horizon, recent studies have explored adaptive action chunking to adjust the action horizon according to the current state or action prediction
\cite{liu2025bidirectional,liang2026adaptive,wang2026vlaknows,nie2026pace}.

Some approaches train additional horizon predictors or horizon-specific policies to learn suitable action horizons
\cite{zhao2026horizon,jing2025mixture},
but require additional data, supervision, or optimization.
Training-free methods instead mainly rely on signals available during inference
\cite{liu2025bidirectional,liang2026adaptive,wang2026vlaknows,nie2026pace}.
For example, predictive uncertainty can be estimated from statistics over stochastic action predictions and used to guide action horizon selection
\cite{liang2026adaptive}.
While informative, such signals provide relatively indirect estimates of the reliability of candidate action horizons.
GeoAAC instead exploits the geometry inherent in the Flow Matching action-generation process, providing a process-level reliability signal for action horizon selection.

\subsection{Flow Matching and Denoising Geometry}

Flow Matching has been widely adopted for continuous action generation in VLA policies, where a continuous velocity field progressively transports initial noise toward robot actions
\cite{black2025pi0,black2025pi05,bjorck2025gr00t,lipman2023flow}.
It has also been explored for efficient and continuous robot control
\cite{jiang2025streaming,black2025realtime}.

Prior work has studied the structure of Flow Matching trajectories from several perspectives.
Rectified Flow analyzes the geometry and straightness of transport trajectories
\cite{liu2023rectifiedflow},
while other studies investigate how coupling strategies affect trajectory structure and sampling behavior
\cite{pooladian2023multisample}.
More recent work has linked denoising-trajectory geometry to predictive uncertainty
\cite{rao2026geometry}.
Building on these observations, we study denoising-trajectory geometry across action prefixes in flow-based VLA policies and use its prefix-wise structure to guide adaptive action horizon selection.

\section{METHODS}

\subsection{Preliminaries: Flow Matching and Prefix-Wise Geometry}
\label{sec:flow_action_generation}

For Vision--Language--Action (VLA) policies equipped with a Flow Matching action head, action chunks are generated by transporting samples from a noise distribution toward the conditional action distribution through a time-dependent vector field \cite{lipman2023flow}. Given the current observation $o$, the action state $\mathbf{x}_{\tau}$ evolves over flow time $\tau \in [0,1]$ according to
\begin{equation}
\frac{d\mathbf{x}_{\tau}}{d\tau}
=
\mathbf{v}_{\theta}(\mathbf{x}_{\tau},\tau \mid o),
\label{eq:flow_ode}
\end{equation}
where $\mathbf{v}_{\theta}$ denotes the learned conditional velocity field. Integrating Eq.~(\ref{eq:flow_ode}) produces an action chunk
$\mathbf{A}=(\mathbf{a}_1,\mathbf{a}_2,\ldots,\mathbf{a}_H)$,
where $H$ denotes the prediction horizon. During numerical integration, the Flow Matching action head evaluates the velocity field at a sequence of integration steps, yielding intermediate velocity predictions
$\{\mathbf{v}^{(t)}\}_{t=0}^{T-1}$
that characterize the evolution of the denoising trajectory.

For adaptive action chunking, each action prefix
$\mathbf{A}_{1:k}=(\mathbf{a}_1,\ldots,\mathbf{a}_k)$,
$k\in\{1,\ldots,H\}$,
corresponds to a candidate action horizon. The intermediate velocity predictions can therefore be restricted to the same prefix, denoted by
$\mathbf{v}_{1:k}^{(t)}$.
This provides a prefix-wise representation of the denoising trajectory across candidate action horizons.

Prior work has linked geometric properties of Flow Matching denoising trajectories to predictive uncertainty
\cite{rao2026geometry}.
We examine whether this relationship also holds across action prefixes, since adaptive action horizon selection requires reliability information for different candidate action horizons.
To this end, we consider an uncorrected prefix-wise geometric measure $U_k$, formally defined in Eq.~(\ref{eq:raw_prefix_geometry}), and compare it with a reference predictive uncertainty $R_k$.

\begin{figure}[t]
    \centering
    \includegraphics[width=\columnwidth]{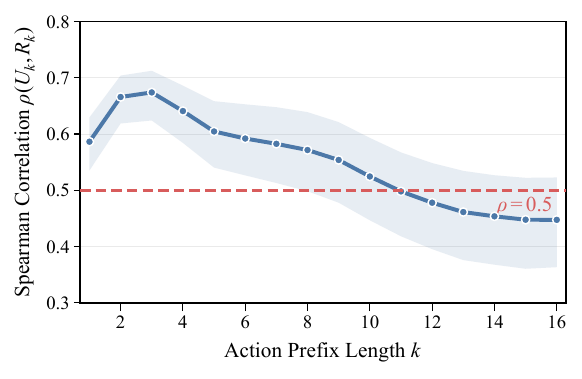}
    \caption{
    \textbf{Correlation between prefix-wise denoising geometry and reference predictive uncertainty.}
    Prefix-wise denoising geometry remains positively correlated with reference predictive uncertainty across action prefix lengths.
    Shaded regions denote 95\% confidence intervals, and the red dashed line indicates $\rho=0.5$ for reference.
    }
    \label{fig:prefix_accel_proxy}
\end{figure}

Using GR00T N1.5, we collect 200 observations from the four LIBERO suites. For each observation, we independently generate 32 stochastic action chunks with $H=16$. The leave-one-out prefix variance obtained from repeated sampling serves as the reference uncertainty $R_k$. As shown in Fig.~\ref{fig:prefix_accel_proxy}, $U_k$ remains positively correlated with $R_k$ across action prefix lengths. This result indicates that the relationship between denoising-trajectory geometry and predictive uncertainty persists at the prefix level, providing empirical support for using prefix-wise geometry to characterize the reliability of candidate action horizons.

\subsection{Geometric Features of the Denoising Trajectory}
\label{sec:denoising_geometry}

Building on the prefix-wise geometric relationship established above, we construct a horizon-wise geometric profile from the intermediate velocity predictions produced during Flow Matching inference.
For an action prefix of length $k$, let $\tau_t$ denote the flow time at the $t$-th integration step and
$\Delta\tau_t = |\tau_{t+1}-\tau_t|$.
We characterize the local geometry of the denoising trajectory through the change between consecutive velocity predictions:
\begin{equation}
b_k^{(t)}
=
\left\|
\mathrm{vec}
\left(
\frac{
\mathbf{v}_{1:k}^{(t+1)}
-
\mathbf{v}_{1:k}^{(t)}
}{
\Delta\tau_t+\epsilon
}
\right)
\right\|_2,
\qquad
t=0,\ldots,T-2,
\label{eq:local_geometry}
\end{equation}
where $\epsilon>0$ is a small constant for numerical stability.
Here, $\mathbf{v}_{1:k}^{(t)}$ denotes the velocity prediction for the first $k$ actions at the $t$-th denoising step.
The difference between consecutive velocity predictions captures local geometric variation along flow time, while normalization by $\Delta\tau_t$ accounts for non-uniform integration intervals.
Computing $b_k^{(t)}$ across denoising steps and action prefixes retains both temporal variation along the denoising trajectory and prefix-wise variation across candidate action horizons.

The uncorrected prefix-wise geometric measure used in Sec.~\ref{sec:flow_action_generation} is obtained by aggregating these local variations:
\begin{equation}
U_k
=
\frac{
T\sum_{t=0}^{T-2} b_k^{(t)}
}{
\sum_{t=0}^{T-1}
\left\|
\mathrm{vec}
\left(
\mathbf{v}_{1:k}^{(t)}
\right)
\right\|_2
+\epsilon
}.
\label{eq:raw_prefix_geometry}
\end{equation}
The factor $T$ normalizes the denominator with respect to the average velocity magnitude across denoising steps.
As shown in Sec.~\ref{sec:flow_action_generation}, $U_k$ remains positively correlated with the reference predictive uncertainty $R_k$ across action prefixes.
However, the reliability of the underlying geometric variations is not uniform across denoising stages.

\begin{figure}[t]
    \centering
    \includegraphics[width=0.9\columnwidth]
    {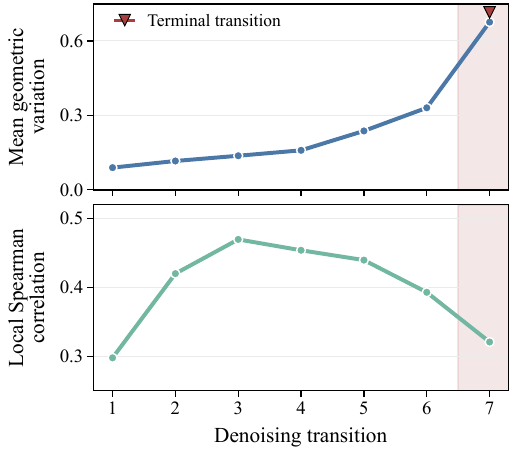}

    \vspace{2pt}
    \centerline{(a) Stage-wise Geometric Reliability}

    \vspace{4pt}

    \includegraphics[width=0.9\columnwidth]
    {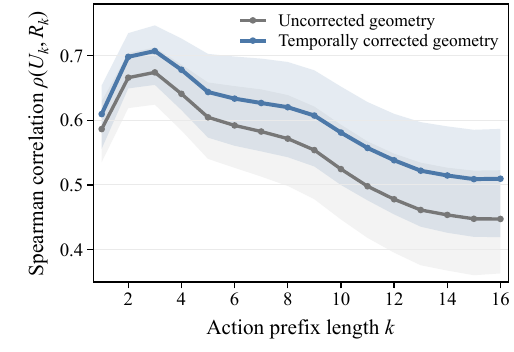}

    \vspace{2pt}
    \centerline{(b) Effect of Temporal Correction}

    \caption{
    \textbf{Temporal correction of denoising geometry.}
    \textbf{(a)} Local geometric variation increases toward late denoising stages, while its correlation with reference uncertainty decreases.
    \textbf{(b)} Temporal correction improves the correlation with reference uncertainty across action prefixes.
    Shaded regions denote 95\% confidence intervals.
    }
    \label{fig:temporal_correction}
\end{figure}

Using the same experimental setup as above, Fig.~\ref{fig:temporal_correction}(a) shows that local geometric variation increases toward later denoising stages, whereas its correlation with the reference uncertainty $R_k$ decreases after the intermediate stages.
Thus, larger geometric variations at later stages do not necessarily provide more reliable uncertainty information.
We therefore account for stage-wise reliability when aggregating the local geometric variations.

Motivated by the reliability pattern in Fig.~\ref{fig:temporal_correction}(a), we introduce a position-dependent temporal weight for each denoising transition.
Let $s_t=t/T$ denote the normalized generation progress at the $t$-th denoising step, and let
$\bar{s}_t=(s_t+s_{t+1})/2$
denote the midpoint of the $t$-th transition.
We define
\begin{equation}
w_t
=
\frac{
\bar{s}_t(1-\bar{s}_t)^2
}{
\frac{1}{T-1}
\sum_{j=0}^{T-2}
\bar{s}_j(1-\bar{s}_j)^2
}.
\label{eq:temporal_weight}
\end{equation}
This weighting function emphasizes intermediate denoising stages while reducing contributions from early and late stages.
The quadratic decay as $s_t$ approaches $1$ suppresses late-stage variations more strongly than early-stage ones, consistent with their lower reliability in Fig.~\ref{fig:temporal_correction}(a).
Mean normalization keeps the average weight equal to one without changing the overall scale.
As shown in Fig.~\ref{fig:temporal_correction}(b), temporal correction improves the correlation between the geometric measure and reference uncertainty across action prefixes.

We then aggregate the temporally corrected local variations and normalize them by the velocity magnitude of the corresponding action prefix:
\begin{equation}
\widetilde{U}_k
=
\frac{
T\sum_{t=0}^{T-2}w_t b_k^{(t)}
}{
\sum_{t=0}^{T-1}
\left\|
\mathrm{vec}
\left(
\mathbf{v}_{1:k}^{(t)}
\right)
\right\|_2
+\epsilon
}.
\label{eq:corrected_prefix_geometry}
\end{equation}
The velocity-magnitude normalization reduces dependence on the overall velocity scale, allowing $\widetilde{U}_k$ to reflect relative geometric variation rather than the magnitude of the velocity predictions.

Applying Eq.~(\ref{eq:corrected_prefix_geometry}) to all candidate action prefixes yields the horizon-wise geometric profile
\begin{equation}
\widetilde{\mathbf U}
=
(\widetilde U_1,\widetilde U_2,\ldots,\widetilde U_H).
\label{eq:horizon_profile}
\end{equation}
This profile describes how denoising geometry evolves as the action prefix expands and serves as the input to adaptive action horizon selection.

\subsection{Adaptive Action Horizon Selection}
\label{sec:horizon_selection}

Given the horizon-wise geometric profile $\widetilde{\mathbf U}$, we determine the execution boundary and thereby the action horizon of the current prediction.
The absolute scale of $\widetilde{\mathbf U}$ can vary across observations, tasks, and generated action chunks, making a fixed threshold difficult to calibrate across settings.
Moreover, local changes in the profile may contain transient fluctuations.
We therefore determine the execution boundary from relative geometric growth and its cumulative distribution across candidate action horizons.

We first measure the relative geometric growth between adjacent action prefixes:
\begin{equation}
r_k
=
\frac{
\widetilde U_k-\widetilde U_{k-1}
}{
|\widetilde U_{k-1}|+\epsilon
},
\qquad
k=2,\ldots,H.
\label{eq:relative_growth}
\end{equation}
Unlike absolute differences, $r_k$ measures the proportional geometric change introduced by extending the action prefix and is therefore less sensitive to variations in signal scale.

We retain only positive geometric growth:
\begin{equation}
e_k
=
\max(r_k,0).
\label{eq:positive_growth}
\end{equation}
A positive $r_k$ indicates increased geometric variation when the candidate action horizon is extended, whereas negative growth provides no evidence for an earlier execution boundary.
Retaining only positive growth also prevents such decreases from canceling boundary evidence accumulated at preceding prefixes.

Rather than selecting the largest local response, which can be dominated by a transient peak, we accumulate the positive geometric growth along the action horizon.
When $\sum_{j=2}^{H}e_j>0$, we define
\begin{equation}
F_k
=
\frac{
\sum_{i=2}^{k}e_i
}{
\sum_{j=2}^{H}e_j
},
\qquad
k=2,\ldots,H,
\label{eq:cumulative_evidence}
\end{equation}
with $F_1=0$.
The resulting $F_k$ forms a normalized cumulative distribution of positive geometric growth.
Isolated responses contribute only a fraction of the total evidence, whereas persistent growth accumulates across neighboring prefixes.

We define the geometry-based execution boundary at the median of this cumulative distribution, i.e., where $F_k$ reaches $0.5$.
This criterion depends on the distribution of relative growth rather than the absolute scale of $\widetilde{\mathbf U}$, avoiding scale-specific threshold calibration while reducing sensitivity to individual local extrema.
When $F_k<0.5<F_{k+1}$, we obtain a continuous boundary estimate by linear interpolation:
\begin{equation}
\hat{k}_{50}
=
k+
\frac{
0.5-F_k
}{
F_{k+1}-F_k
},
\qquad
k=1,\ldots,H-1.
\label{eq:k50_interp}
\end{equation}
Rounding $\hat{k}_{50}$ to the nearest action step gives $k_{50}$.
If $F_k=0.5$, we set $k_{50}=k$.
If no positive geometric growth is observed, the profile provides no evidence for early truncation, and we set $k_{50}=H$.

We further apply a motion-aware lower bound as a safeguard against excessively short action horizons when the predicted motion is small.
In this regime, executing only a few low-magnitude actions may produce little state change while triggering another policy inference.
Let $M(\mathbf A)$ denote the overall motion magnitude of the predicted action chunk, measured from its translational and rotational components, and let $\alpha>0$ denote a fixed reference scale.
We define
\begin{equation}
q
=
\min
\left(
\frac{M(\mathbf A)}{\alpha},
1
\right),
\label{eq:motion_level}
\end{equation}
and the corresponding minimum action horizon as
\begin{equation}
k_{\mathrm{motion}}
=
\left\lfloor
k_{\min}
+
(1-q)(H-k_{\min})
+
\frac{1}{2}
\right\rfloor .
\label{eq:motion_bound}
\end{equation}
Larger predicted motion permits a shorter minimum action horizon, whereas smaller motion retains a longer minimum action horizon.
Here, $k_{\min}$ is a fixed minimum action horizon parameter, and both $\alpha$ and $k_{\min}$ are fixed across tasks.

The final action horizon is
\begin{equation}
k^*
=
\max
\left(
k_{50},
k_{\mathrm{motion}}
\right).
\label{eq:final_horizon}
\end{equation}
The policy executes the first $k^*$ predicted actions before acquiring a new observation and replanning, resulting in adaptive closed-loop execution.

\section{EXPERIMENTS}

\subsection{Experimental Settings}
\label{sec:experimental_settings}

\textbf{Models.} We evaluate two flow-based VLA policies, GR00T N1.5 \cite{bjorck2025gr00t} and $\pi_{0.5}$ \cite{black2025pi05}. For GR00T N1.5, we set the prediction horizon to $H=16$, with 8 denoising steps on LIBERO and 4 on RoboCasa365. For $\pi_{0.5}$, we use a prediction horizon of $H=10$ and 10 denoising steps on LIBERO-Pro.

\textbf{Benchmarks.} We evaluate on LIBERO \cite{liu2023libero}, RoboCasa365 \cite{nasiriany2026robocasa365}, and LIBERO-Pro \cite{zhou2025liberopro}. On LIBERO, we use all 40 tasks from the Spatial, Object, Goal, and LIBERO-10 suites, covering spatial-relation reasoning, object-centric manipulation, goal-conditioned manipulation, and longer-horizon multi-stage tasks, with 50 rollouts per task. On RoboCasa365, we evaluate 18 single-skill manipulation tasks from the \texttt{target/atomic\_seen} split, including pick-and-place, articulated-object manipulation, and appliance interaction, also with 50 rollouts per task. On LIBERO-Pro, we evaluate 10 LIBERO-Object tasks under three levels of position perturbation, Shift-0.2, Shift-0.3, and Shift-0.4, to assess robustness under spatial distribution shifts.

\textbf{Baselines.} We compare with fixed action horizons and two adaptive baselines: multi-sampling-based (MS) \cite{liang2026adaptive}, which selects action horizons from multiple stochastic action predictions, and self-attention-based (SA) \cite{wang2026vlaknows}, which uses internal action self-attention for action horizon selection.

\subsection{Simulation}
\label{sec:simulation_experiments}

%%table1：libero%%
\begin{table}[t]
\caption{
Success rates (\%) on LIBERO under different action chunking strategies with GR00T N1.5 and $\pi_{0.5}$.
}
\label{tab:libero_results}
\centering
\small
\begin{tabular*}{\columnwidth}{@{\extracolsep{\fill}}lccccc@{}}
\toprule
Method & Spatial & Long & Object & Goal & Avg. \\
\midrule
GR00T ($h=2$)
& 95.0 & 79.0 & 93.6 & 93.0 & 90.2 \\
GR00T ($h=4$)
& 95.8 & 82.6 & 97.0 & 94.4 & 92.5 \\
GR00T ($h=8$)
& 95.0 & 88.4 & 97.4 & \textbf{97.8} & 94.7 \\
GR00T ($h=12$)
& 95.6 & 88.2 & 97.2 & 97.0 & 94.5 \\
GR00T ($h=16$)
& 95.0 & 88.2 & 97.0 & 96.0 & 94.1 \\
GR00T+MS
& \textbf{97.2} & 88.0 & 96.6 & 96.4 & 94.6 \\
GR00T+SA
& 96.2 & 87.4 & 95.8 & 95.8 & 93.8 \\
GR00T+GeoAAC
& 96.4 & \textbf{89.2} & \textbf{99.0} & 97.2 & \textbf{95.5} \\
\midrule
$\pi_{0.5}$ ($h=5$)
& 98.5 & 93.2 & \textbf{98.8} & 98.0 & 97.1 \\
$\pi_{0.5}$+MS
& 98.8 & 94.4 & 96.6 & \textbf{98.8} & 97.2 \\
$\pi_{0.5}$+SA
& \textbf{99.0} & 93.2 & 98.0 & 98.2 & 97.1 \\
$\pi_{0.5}$+GeoAAC
& 98.6 & \textbf{96.4} & 98.6 & 98.4 & \textbf{98.0} \\
\bottomrule
\end{tabular*}
\end{table}

%%table 2：libero-pro
\begin{table}[t]
\caption{
Success rates (\%) on LIBERO-Pro under different position-shift settings with $\pi_{0.5}$.
}
\label{tab:liberopro_results}
\centering
\small
\begin{tabular*}{\columnwidth}{@{\extracolsep{\fill}}lcccc@{}}
\toprule
Method & Shift-0.2 & Shift-0.3 & Shift-0.4 & Avg. \\
\midrule
$\pi_{0.5}$ ($h=5$)
& 53.2 & 29.9 & 9.5 & 30.9 \\
$\pi_{0.5}$+SA
& 57.5 & 34.3 & 9.5 & 33.8 \\
$\pi_{0.5}$+MS
& 57.4 & 35.5 & \textbf{12.8} & 35.2 \\
$\pi_{0.5}$+GeoAAC
& \textbf{58.5} & \textbf{37.6} & 12.4 & \textbf{36.2} \\
\bottomrule
\end{tabular*}
\end{table}

%%table-three Robocasa 提前放
\begin{table*}[!t]
\caption{
Success rates (\%) on RoboCasa365 under different action chunking strategies with GR00T N1.5.
}
\label{tab:robocasa_results}
\centering
\small

\begin{tabular*}{\textwidth}{@{\extracolsep{\fill}}lcccccccc@{}}
\toprule
Task & GeoAAC & Fixed-2 & Fixed-4 & Fixed-8 & Fixed-12 & Fixed-16 & MS & SA \\
\midrule

CloseBlenderLid
& 36.0 & 8.0 & 14.0 & 28.0 & 36.0 & \textbf{40.0} & 34.0 & 38.0 \\

CloseFridge
& 80.0 & 38.0 & 42.0 & 56.0 & 68.0 & 78.0 & 58.0 & \textbf{84.0} \\

CloseToasterOvenDoor
& \textbf{90.0} & 44.0 & 56.0 & 78.0 & \textbf{90.0} & 82.0 & 80.0 & 86.0 \\

CoffeeSetupMug
& 70.0 & 20.0 & 34.0 & 62.0 & 62.0 & 70.0 & 60.0 & \textbf{72.0} \\

NavigateKitchen
& \textbf{68.0} & 16.0 & 26.0 & 40.0 & 50.0 & 62.0 & 64.0 & 42.0 \\

OpenCabinet
& 92.0 & 50.0 & 72.0 & 88.0 & 92.0 & 84.0 & 92.0 & \textbf{94.0} \\

OpenDrawer
& \textbf{94.0} & 20.0 & 32.0 & 60.0 & 76.0 & 70.0 & 78.0 & 82.0 \\

OpenStandMixerHead
& 92.0 & 62.0 & 80.0 & 86.0 & 94.0 & \textbf{96.0} & 94.0 & 94.0 \\

PickPlaceCounterToCabinet
& \textbf{68.0} & 48.0 & 46.0 & 52.0 & 54.0 & 60.0 & \textbf{68.0} & \textbf{68.0} \\

PickPlaceCounterToStove
& \textbf{84.0} & 48.0 & 62.0 & 78.0 & 74.0 & 72.0 & 82.0 & 76.0 \\

PickPlaceDrawerToCounter
& 54.0 & 10.0 & 22.0 & 44.0 & 32.0 & 38.0 & \textbf{58.0} & 46.0 \\

PickPlaceSinkToCounter
& 76.0 & 36.0 & 64.0 & 56.0 & 80.0 & 72.0 & \textbf{84.0} & 76.0 \\

PickPlaceToasterToCounter
& 72.0 & 24.0 & 46.0 & 68.0 & 84.0 & 62.0 & \textbf{86.0} & 70.0 \\

SlideDishwasherRack
& \textbf{80.0} & 50.0 & 56.0 & 68.0 & 68.0 & 78.0 & \textbf{80.0} & 74.0 \\

TurnOffStove
& \textbf{56.0} & 10.0 & 24.0 & 30.0 & 48.0 & 46.0 & 24.0 & 42.0 \\

TurnOnElectricKettle
& \textbf{90.0} & 34.0 & 48.0 & 62.0 & 62.0 & 72.0 & 78.0 & 82.0 \\

TurnOnMicrowave
& \textbf{78.0} & 12.0 & 26.0 & 60.0 & 64.0 & 60.0 & \textbf{78.0} & 60.0 \\

TurnOnSinkFaucet
& 72.0 & 12.0 & 34.0 & 40.0 & 62.0 & 50.0 & \textbf{82.0} & 54.0 \\

\midrule
Overall
& \textbf{75.1} & 30.1 & 43.6 & 58.7 & 66.4 & 66.2 & 71.1 & 68.9 \\

\bottomrule
\end{tabular*}
\end{table*}

\textbf{LIBERO.}
Table~\ref{tab:libero_results} reports the success rates of different action chunking strategies on LIBERO. With GR00T N1.5, GeoAAC achieves the highest average success rate of 95.5\%, outperforming the best fixed-action-horizon baseline, fixed-$8$ (94.7\%), as well as MS (94.6\%) and SA (93.8\%). GeoAAC also achieves the best performance on the Object and Long suites, reaching 99.0\% and 89.2\%, respectively. These results show that adaptive action horizon selection improves the overall task performance of GR00T N1.5.

\begin{figure}[t]
    \centering
    \makebox[\linewidth][c]{%
        \includegraphics[width=1.08\linewidth]{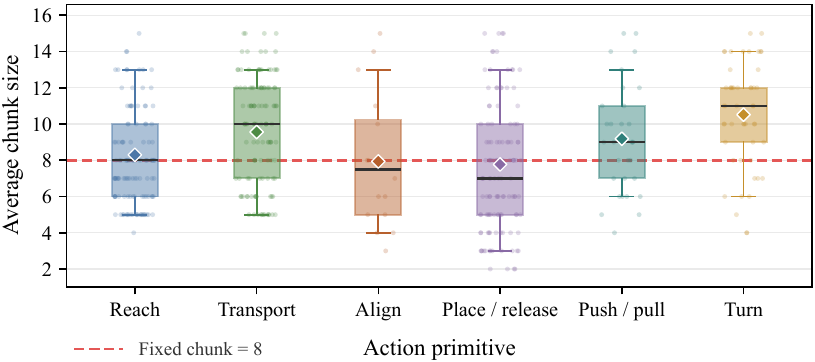}
    }
    \caption{
    \textbf{Adaptive action horizon selection across action primitives.}
    Each point denotes the average action horizon of one action primitive within an episode, and the dashed line marks fixed-$8$, the best fixed-action-horizon baseline. Different action primitives exhibit distinct action-horizon preferences: continuous motions favor longer horizons, while goal-proximal alignment and placement favor shorter horizons.
    }
    \label{fig:primitive_chunk}
\end{figure}

To analyze how GeoAAC responds to different action primitives, we examine the action horizons selected for different local motion patterns. As shown in Fig.~\ref{fig:primitive_chunk}, Align and Place/release, which require relatively precise pose control and frequent closed-loop correction, have shorter average action horizons of 7.94 and 7.77, respectively. In contrast, Transport, Push/pull, and Turn exhibit longer average action horizons of 9.56, 9.19, and 10.52, respectively, corresponding to more continuous translational motion, contact-constrained motion, and rotational manipulation. Reach has an average action horizon of 8.30, close to the fixed-$8$ baseline. These results show that GeoAAC does not apply a uniform action horizon across action primitives, but instead adapts the action horizon to local motion characteristics, using shorter horizons for alignment and placement while maintaining longer horizons for continuous transport, contact motion, and rotation.

We further evaluate GeoAAC with $\pi_{0.5}$ to examine its applicability across different flow-based VLA policies. GeoAAC achieves an average success rate of 98.0\%, compared with 97.1\% for fixed-$5$ and 97.2\% for MS. On the Long suite, GeoAAC improves the success rate from 93.2\% with fixed-$5$ to 96.4\%. These results show that the performance advantage of GeoAAC generalizes across different flow-based VLA policies.

\textbf{LIBERO-Pro.}
We further evaluate GeoAAC on LIBERO-Pro \cite{zhou2025liberopro} under three position-perturbation settings: Shift-0.2, Shift-0.3, and Shift-0.4.

Table~\ref{tab:liberopro_results} reports the success rates on LIBERO-Pro. GeoAAC achieves the highest average success rate of 36.2\%, outperforming the fixed-action-horizon baseline by 5.3 percentage points and also surpassing the self-attention-based (SA) \cite{wang2026vlaknows} and multi-sampling-based (MS) \cite{liang2026adaptive} adaptive baselines. Under Shift-0.2 and Shift-0.3, GeoAAC reaches 58.5\% and 37.6\%, respectively, while remaining competitive under the stronger Shift-0.4 perturbation. These results show that GeoAAC maintains its advantage under changes in object-position distribution by adapting the action horizon to the current prediction.

\textbf{RoboCasa365.}
We further evaluate GeoAAC on RoboCasa365 \cite{nasiriany2026robocasa365} to assess its performance under more diverse household manipulation scenarios. Table~\ref{tab:robocasa_results} reports the success rates of different action chunking strategies.

With GR00T N1.5 \cite{bjorck2025gr00t}, GeoAAC achieves the highest average success rate of 75.1\%. It outperforms the best fixed-action-horizon baseline by 8.7 percentage points and exceeds the adaptive baselines MS \cite{liang2026adaptive} and SA \cite{wang2026vlaknows} by 4.0 and 6.2 percentage points, respectively. GeoAAC also achieves or ties the best performance on 9 of the 18 tasks. These results show that GeoAAC maintains a consistent performance advantage across diverse household manipulation scenarios.

\subsection{Real-World Experiments}

\textbf{Setup and Tasks.}
We further evaluate GeoAAC on a real-world dual-arm manipulation platform consisting of two PiperX robotic arms and three Intel RealSense D435i cameras, providing one third-person view and two wrist-mounted views.
The base policy is a flow-based VLA policy built on a Qwen3-VL-2B-Instruct vision-language backbone with a Flow Matching action expert.
At each inference, the policy predicts an action chunk with a prediction horizon of $H=50$, while the robot is controlled at 50~Hz.

\begin{figure}[t]
    \centering
    \includegraphics[width=\columnwidth]{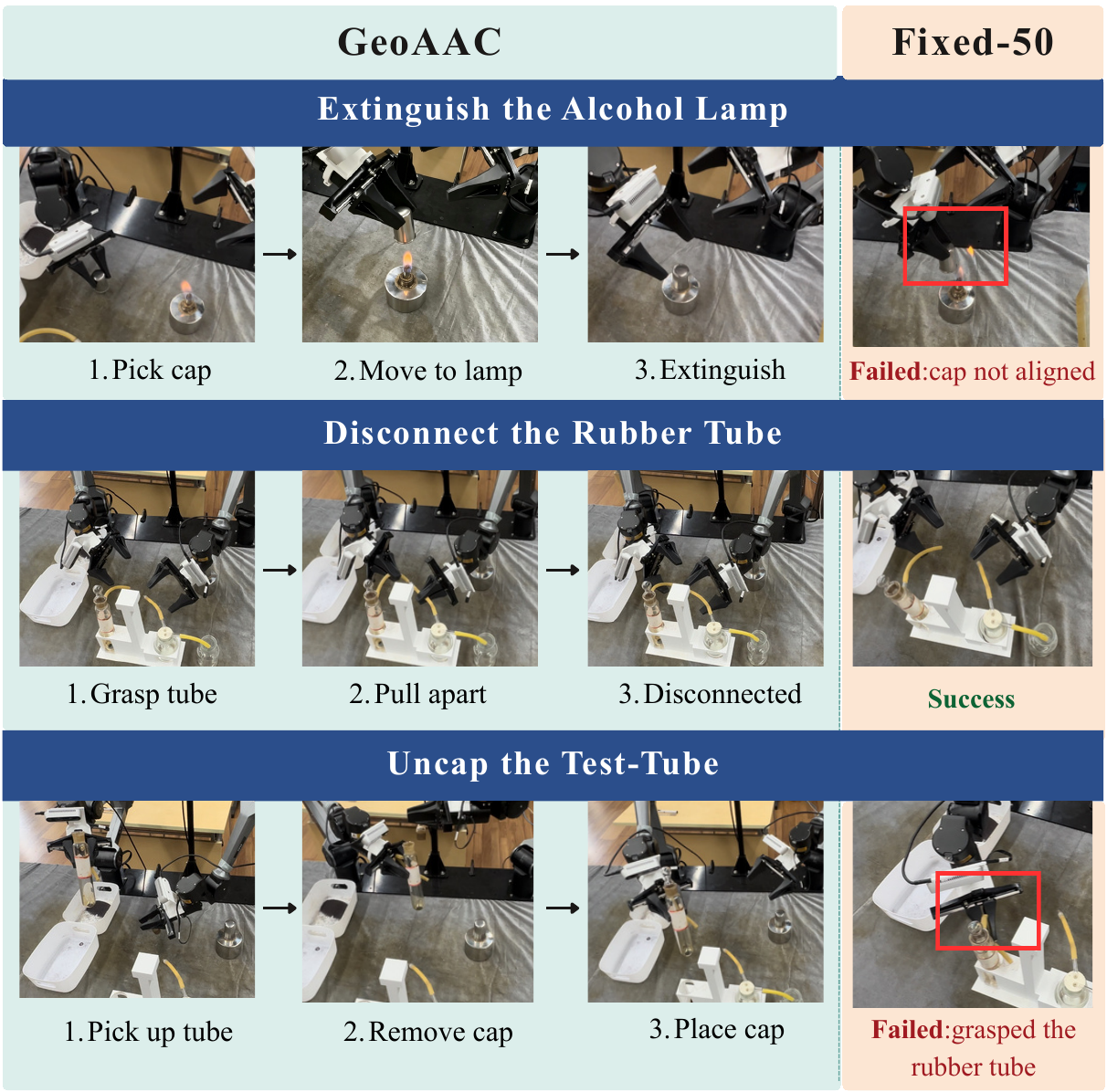}
    \caption{
    \textbf{Real-world manipulation experiments.}
    GeoAAC and Fixed-$50$ are evaluated on alcohol lamp extinguishing, rubber tube disconnection, and test-tube uncapping.
    Fixed-$50$ fails due to cap misalignment in the first task and incorrect object grasping in the third task, while successfully completing rubber tube disconnection.
    }
    \label{fig:real_tasks}
\end{figure}

As shown in Fig.~\ref{fig:real_tasks}, we consider three manipulation tasks with different interaction characteristics: Alcohol Lamp Extinguishing, Rubber Tube Disconnection, and Test-Tube Uncapping.
We collect 50 teleoperated demonstrations per task, resulting in 150 real-world trajectories for training.
Fixed-$50$ and GeoAAC use the same policy checkpoint and differ only in their action horizon.
Fixed-$50$ executes the full $50$-step action chunk, whereas GeoAAC adaptively determines the execution boundary using the denoising-trajectory geometry from the same Flow Matching generation.
Each method is evaluated over 30 trials per task, resulting in 180 trials in total.

\begin{table}[t]
\caption{
Success rates (\%) on real-world manipulation tasks.
Each method is evaluated over 30 trials per task.
}
\label{tab:real_world_results}
\centering
\small
\begin{tabular*}{\columnwidth}{@{\extracolsep{\fill}}lcccc@{}}
\toprule
Method & Extinguish & Disconnect & Uncap & Avg. \\
\midrule
Fixed-$50$
& 16.7 & 76.7 & 66.7 & 53.3 \\
GeoAAC
& \textbf{36.7} & \textbf{100.0} & \textbf{86.7} & \textbf{74.4} \\
\bottomrule
\end{tabular*}
\end{table}

\textbf{Results.}
As shown in Table~\ref{tab:real_world_results}, GeoAAC improves the success rate across all three tasks, increasing the average success rate from 53.3\% to 74.4\%, an absolute improvement of 21.1 percentage points.
On Rubber Tube Disconnection, GeoAAC achieves 100.0\% success, compared with 76.7\% for Fixed-$50$.
Since both settings use the same policy parameters and training data, these results show that adapting the action horizon alone improves overall real-world performance.

Qualitative observations further highlight the benefit of adaptive action horizon selection during stages requiring precise closed-loop adjustment.
In Test-Tube Uncapping, Fixed-$50$ is more prone to empty grasps or grasping the wrong target, whereas GeoAAC acquires updated observations more frequently during grasping and uncapping.
In Rubber Tube Disconnection, GeoAAC also adjusts the gripper and wrist pose when approaching the target, making execution less sensitive to variations in the initial robot configuration.
These observations are consistent with the varying need for closed-loop feedback across manipulation stages.

\section{Conclusion}

We presented \textbf{GeoAAC}, a geometry-based adaptive action chunking method for flow-based VLA policies. GeoAAC exploits the prefix-wise geometry of Flow Matching denoising trajectories from a single action generation process and constructs a horizon-wise geometric profile to characterize reliability-related variations along the predicted action sequence. Based on the relative growth and cumulative distribution of this profile, GeoAAC adaptively determines the execution boundary without additional training. Experiments with GR00T N1.5 and $\pi_{0.5}$ across LIBERO, LIBERO-Pro, and RoboCasa365 demonstrate consistent improvements over fixed-action-horizon and adaptive baselines. Real-world experiments further validate the effectiveness of GeoAAC, improving the average success rate from 53.3\% with Fixed-$50$ to 74.4\% across three manipulation tasks.

Our current approach focuses on geometric information available within the Flow Matching action-generation process for action horizon adaptation. Future work could further investigate when task-relevant information should be acquired and incorporated during robot execution, extending adaptive action horizon selection toward more flexible closed-loop interaction.


%%%%%%%%%%%%%%%%%%%%%%%%%%%%%%%%%%%%%%%%%%%%%%%%%%%%%%%%%%%%%%%%%%%%%%%%%%%%%%%%

%%%%%%%%%%%%%%%%%%%%%%%%%%%%%%%%%%%%%%%%%%%%%%%%%%%%%%%%%%%%%%%%%%%%%%%%%%%%%%%%

%%%%%%%%%%%%%%%%%%%%%%%%%%%%%%%%%%%%%%%%%%%%%%%%%%%%%%%%%%%%%%%%%%%%%%%%%%%%%%%%
%%\section*{APPENDIX}

%%\section*{ACKNOWLEDGMENT}

%%%%%%%%%%%%%%%%%%%%%%%%%%%%%%%%%%%%%%%%%%%%%%%%%%%%%%%%%%%%%%%%%%%%%%%%%%%%%%%%

\bibliographystyle{IEEEtran}
\bibliography{references}

\end{document}